\documentclass[conference]{IEEEtran}
\IEEEoverridecommandlockouts

\usepackage[letterpaper]{geometry}
\newgeometry{left=1.57cm,right=1.57cm,bottom=2.2cm,top=1.8cm}

\usepackage{setspace}
\usepackage{booktabs} 
\usepackage[noadjust]{cite}
\usepackage{url}
\usepackage{amsmath}
\usepackage{amssymb}
\usepackage{amsfonts}
\usepackage{float}
\usepackage{graphicx}
\usepackage{pifont}
\usepackage{xcolor}
\usepackage{colortbl}
\usepackage{multirow}
\usepackage{mathtools}
\usepackage{algorithm}
\usepackage{algpseudocode}
\usepackage{tabto}
\usepackage{todonotes}
\usepackage{hhline}
\usepackage{enumitem}

\usepackage[belowskip=-4pt,aboveskip=4pt,font=small,labelfont={it}]{caption}

\def\BibTeX{{\rm B\kern-.05em{\sc i\kern-.025em b}\kern-.08em
    T\kern-.1667em\lower.7ex\hbox{E}\kern-.125emX}}
        \makeatletter
\renewcommand\footnoterule{%
  \kern-3\p@
  \hrule\@width 0.5\columnwidth
  \kern2.6\p@}
  \makeatother
  
\usepackage{fancyhdr,lipsum}
\fancypagestyle{firstpage}{
  \fancyhf{}
  \fancyhead[C]{To appear at the 23rd Design, Automation and Test in Europe (DATE 2020). Grenoble, France.}
  \fancyfoot[C]{\thepage}
}
\begin{document}
\font\myfont=cmr12 at 22pt
\setlength\parskip{1pt}
\setlength{\abovedisplayskip}{3pt}
\setlength{\belowdisplayskip}{3pt}
\setlength{\textfloatsep}{3pt}
\setlength{\floatsep}{3pt}
\title{\myfont FANNet: Formal Analysis of Noise Tolerance, Training Bias and Input Sensitivity in Neural Networks} 
\author{\IEEEauthorblockN{Mahum Naseer\textsuperscript{ 1*}\thanks{*Mahum Naseer and Mishal Fatima Minhas both are lead authors, and have equal contributions.}, Mishal Fatima Minhas\textsuperscript{ 2*}, Faiq Khalid\textsuperscript{ 1}, Muhammad Abdullah Hanif\textsuperscript{ 1},\\ Osman Hasan\textsuperscript{ 2}, Muhammad Shafique\textsuperscript{ 1}}
   	\IEEEauthorblockA{\textsuperscript{ 1}\textit{Technische Universit\"at Wien (TU Wien), Vienna, Austria}\\
  	\textsuperscript{ 2}\textit{National University of Sciences and Technology (NUST), Islamabad, Pakistan}\\
  	\{mahum.naseer,faiq.khalid,muhammad.hanif,muhammad.shafique\}@tuwien.ac.at,\\ \{mminhas.msee15seecs, osman.hasan\}@nust.seecs.edu.pk}
 }
 
\maketitle
\thispagestyle{firstpage}

\begin{abstract}
With a constant improvement in the network architectures and training methodologies, Neural Networks (NNs) are increasingly being deployed in real-world Machine Learning systems. However, despite their impressive performance on ``known inputs", these NNs can fail absurdly on the ``unseen inputs", especially if these real-time inputs deviate from the training dataset distributions, or contain certain types of input noise. This indicates the low noise tolerance of NNs, which is a major reason for the recent increase of adversarial attacks. This is a serious concern, particularly for safety-critical applications, where inaccurate results lead to dire consequences. We propose a novel methodology that leverages model checking for the Formal Analysis of Neural Network (FANNet) under different input noise ranges. Our methodology allows us to rigorously analyze the noise tolerance of NNs, their input node sensitivity, and the effects of training bias on their performance, e.g., in terms of classification accuracy. For evaluation, we use a feed-forward fully-connected NN architecture trained for the Leukemia classification. Our experimental results show $\pm 11\%$ noise tolerance for the given trained network, identify the most sensitive input nodes, and confirm the biasness of the available training dataset.
\end{abstract}
\begin{IEEEkeywords}
Neural Networks, Formal Methods, Model Checking, Formal Analysis, Adversarial Machine Learning.
\end{IEEEkeywords}


\vspace{-1mm}
\section{Introduction} \label{Intro}
\vspace{-1mm}
Due to their remarkable learning capacity and astounding accuracy on labelled datasets, Neural Networks (NNs) have become a ubiquitous paradigm in the Machine Learning (ML) based smart systems. Several present day applications including object detectors, speech recognizers, malware detectors, and even the safety-critical applications like medical diagnosis and autonomous driving increasingly rely on NNs~\cite{healthcare}\cite{auto-drive}. However, NNs are extremely sensitive to changes in input, and even the imperceptible input noise can cause the NN to misclassify~\cite{AdvAttack}. This makes obtaining reliable guarantees regarding correct NN behavior a significant need.

A standard practice to ensure a good performance of a trained NN is to test it with a given validation dataset. However, the applications where NNs are deployed often have infinite input domain, for instance, when subjected to lifelong learning scenarios \cite{L2Mdarpa}. \textit{This makes exhaustive testing infeasible for NNs}. Moreover, the validation datasets are insufficient to generalize to the vast input domain. This undermines testing as an option to obtain reliable guarantees regarding the NN.

Intuitively, since formal methods ought to provide reliable guarantees regarding a system's behavior, there has been an increasing interest to use formal methods \cite{reluplex}\cite{MIT} to ascertain correctness of the trained NNs. However, the non-linearity and non-convexity of the NNs, along with the NP-completeness \cite{reluplex} of verifying even the piece-wise linear NNs, makes formal analysis of NNs an extremely challenging task. The current objective of formal analysis of NNs is to ensure the robustness of a trained network in the presence of small input perturbations. The idea is to ensure that adding perturbations, known as the \textit{adversarial noise}, to the network input must not change output classification of the network.
\begin{figure}[!t]
	\centering
	\includegraphics[width=\linewidth]{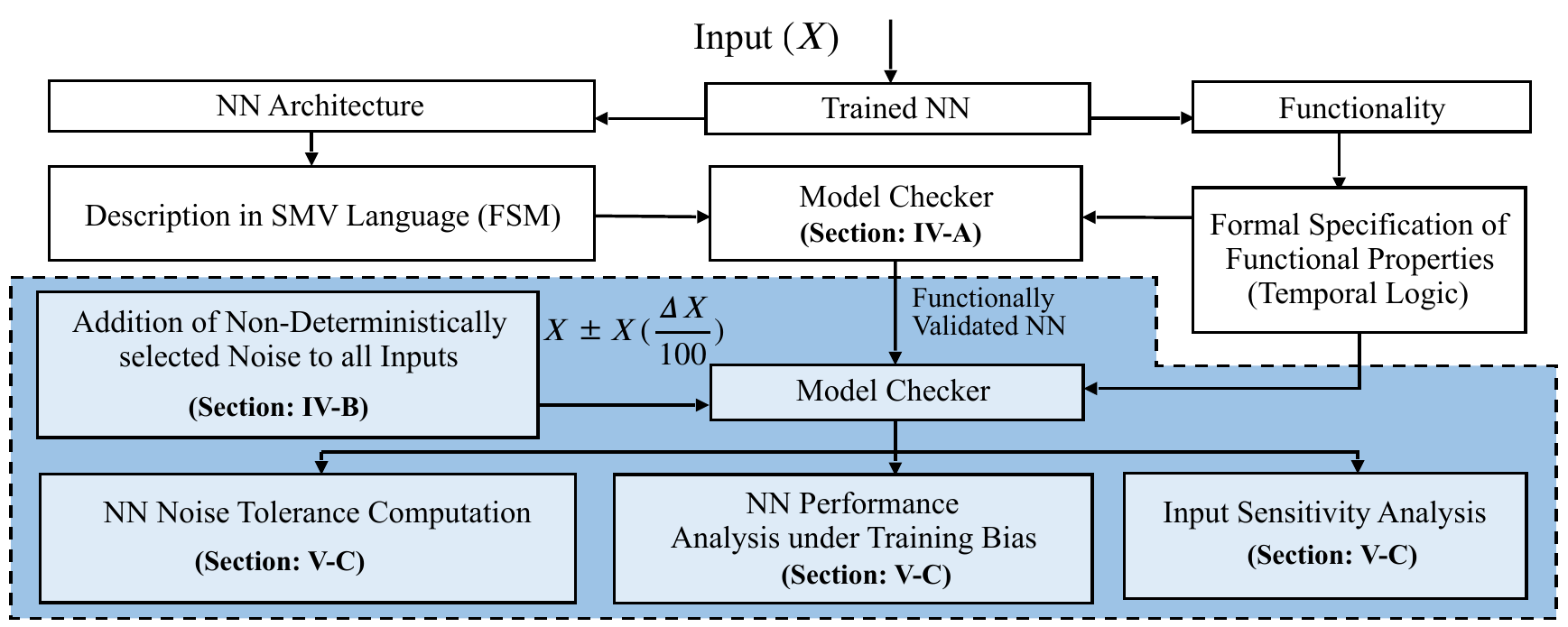}
	\caption{Overview of FANNet with Novel Contributions for the Formal Analysis of NN properties, by exploiting Model Checking}
	\label{novel}
\end{figure}
\subsection{Limitations in State-of-the-Art and Open Challenges}
There remains several unresolved problems in the state-of-the-art regarding formal analysis of NNs  (which will be discussed further in Sections \ref{RW}):
\begin{enumerate}[leftmargin=*]
    \item The current formal verification frameworks like \cite{vmware}\cite{Z3} mostly focus on ensuring network robustness in the presence of input noise. \textit{However, they seldom focus on obtaining the noise tolerance of the trained NN}. 
    \item The decision making of the NN is often incomprehensible to humans. \textit{The understanding regarding the sensitivity of the individual NN input nodes, which often lacks in current works, may provide useful insights to the NN inference}.
    \item The effects of training bias of a NN deployed in practical systems are well-studied. \textit{However, its effects under input noise and use formal analysis to ensure fair training procedure remain scarcely studied topics}.
\end{enumerate}

\subsection{Our Novel Contributions and Concept Overview}
We propose a formal analysis methodology, FANNet, to address the aforementioned challenges
; see Fig. \ref{novel}. The novel contributions of this work (elaborated in Sections \ref{PM}, \ref{Result}) are:
\begin{enumerate}[leftmargin=*]
    \item Formal modeling and analysis methodology to analyze the trained NN as a state-space model, using model checking.
    \item Providing realistic estimates for the network tolerance in the presence of adversarial noise. 
    \item Studying input node sensitivity in the presence of noise.
    \item Analyzing the effects of training bias on network accuracy.
    \item Performing a case study on Leukemia Detection to demonstrate the practical significance of the above analysis.
\end{enumerate}
\section{Related Work} \label{RW}
\textbf{Verification for NN Robustness} - Verification for ensuring \textit{network robustness}, has been an active domain of research for NNs. These works focus on either: (a) expressing the network and its robustness property in Conjunctive Normal Form (CNF), and verifying it using a satisfiability (SAT) solver \cite{reluplex}\cite{vmware,Z3,BNNfortiss}, (b) transforming the problem into a set of linear constraints and objective function, and verifying the network as an optimization problem using a linear programming (LP) solver \cite{MIT}\cite{dutta2018,imperial2018,wang2018columbia}, or (c) using a combination of both SAT and LP solvers \cite{ehlers}, under a specified $L^p-norm$ space around the seed inputs. Unlike the current literature, we analyze NN properties beyond network robustness, using model checking.

\textbf{Training Bias and Input Node Sensitivity} - \textit{Biased training} data is among the leading causes for a biased NN \cite{BIASsurvey}. To solve the problem, either specialized training algorithms are proposed  \cite{FairTraining1}\cite{FairTraining2}, or formal analysis is used to ensure \textit{fairness} of the NN \cite{fairness-verification} in presence of larger (observable) bias.
On the other hand, the literature focusing on \textit{input node sensitivity} of NN \cite{input-sens1}\cite{input-sens2} generally aims either to identify input features that can be pruned out of the network without hampering the network performance, in the absence of noise, or use input noise to study nodes' response for selection of input features \cite{IS2002}\cite{IS2017}. 
Our work, on the other hand, takes a more qualitative approach with input sensitivity, and identifies the input nodes that require precise input acquisition to avoid misclassification.
\section{Preliminaries} \label{Pre}
This section provides the background knowledge on basic NN concepts and properties necessary to understand our novel contributions. The notations introduced in this section will be followed in the subsequent sections.
\vspace{-1mm}
\subsection{Neural Network and its Properties} \label{Pre-NN}
We consider feed-forward network, with $L$ fully-connected (FC) layers, each containing $N$ neurons. Each neuron of a layer is connected to all neurons of the previous layer with a deterministic relation between inputs, the associated weight matrix $w$ and the bias vector $b$. The overall network is given by $f(x): x \rightarrow x^{L-1}$, where $x$ and $x^{L-1}$ represent input and output layers of the network respectively. 


\begin{figure*}
	\centering
	\includegraphics[width=\linewidth]{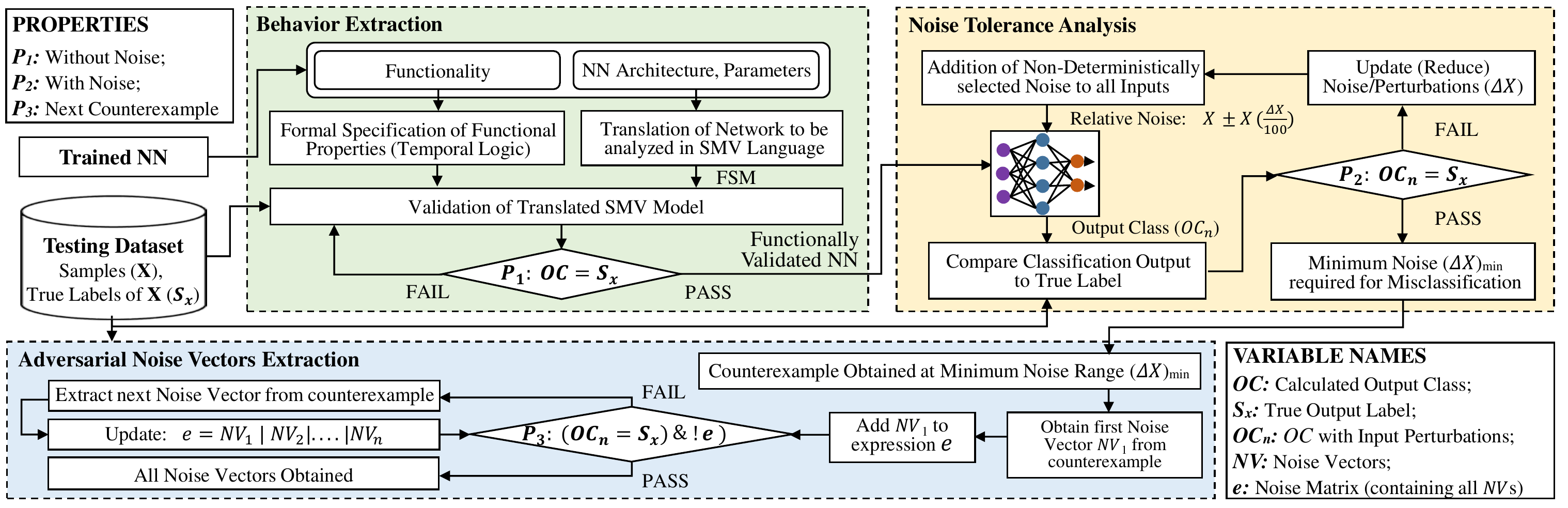}
	\vspace*{-5mm}
	\caption{FANNet: A Methodology for Formally Analyzing Neural Network Properties using Model Checking}
	\vspace{-10pt}
	\label{meth}
\end{figure*} 
Except the input layer, each layer is associated with a non-linear activation function $\sigma$. The most common activation functions are: Rectified Linear Unit (ReLU), maxpool, sigmoid and softmax. In this paper, we focus on ReLU and maxpool activations due to their predominant use in practical NNs recently \cite{densenet}.

\textbf{Robustness} is an essential property of a NN that determines its sturdiness in the presence of small noise $\Delta x$. A network is said to be robust if the addition of $\Delta x$ to the input $x$ does not change the predicted output label:
\begin{equation}
    x'=x+\Delta x: ~~~~~|x-x'|<\epsilon \implies f(x') = f(x)
\end{equation}

\textbf{Noise Tolerance} defines the \textit{minimum} noise $(\Delta x)_{min}$ added to the  input $x$,  such that the addition of any further noise causes the network to misclassify the input. Alternatively, it determines the \textit{maximum} noise $(\Delta x)_{max}$ that could be added to the input $x$, for which the network output equals the \textit{true} label for the input $x$, i.e., $S_x$: 
\begin{equation}
    x'=x+(\Delta x)_{max}: f(x') = S_x
\end{equation}

\textbf{Input Node Sensitivity} indicates the \textit{robustness} of individual input nodes to the input perturbations. The neurons (or nodes) in the input layer typically represent distinct, although often correlated, features of the input. 
An input node $x^0_a$ is said to be insensitive to a small noise $\alpha$, if the addition of this noise does not change the predicted output label:
\begin{equation}
    \begin{split}
        &f((x^0_a + \alpha), (x^0_b)) = f(x^0), ~~~~\\ 
        &x^0 = [x^0_1,...,x^0_N],~ x^0_b = x^0\setminus x^0_a,~ a \in [1,2,...,N],
    \end{split}
\end{equation}

\textbf{Training Bias} is generally the result of a biased training dataset, i.e., a dataset with more samples pointing to a certain output label, which often retains even for the unseen inputs. Formally, it can be expressed as a network property, where the addition of noise $\Delta x$ changes the output label for an input $X_1$, but not for input $X_2$:
\begin{equation}
    f(X_1+\Delta x) \neq S_{X_1} \land f(X_2+\Delta x) = S_{X_2}
\end{equation}

\subsection{Model Checking using nuXmv} \label{Pre-MC}
Model checking is a formal analysis technique that enables rigorous verification of the system model defined as a state-space. The standard procedure used during model checking includes the representation of system and its properties in the formal language of the model checker. Model checker then searches the system's state-space either to ensure functional correctness or to find counterexamples in case of failure of the desired properties. The model checkers can either be based on \textit{Satisfiability solvers (SAT/SMT)} or \textit{Binary Decision Diagrams (BDDs)}. BDD-based model checkers are limited by space due to their \texttt{PSPACE}-complete computational complexity. On the other hand, SAT-based model checkers, despite their \texttt{NP}-complete computational complexity, can handle larger number of variables. As will be shown in Section \ref{Result}, the state-space of NN, in the presence of noise, can grow exponentially. Hence, the SMT-based symbolic model checker, \textit{nuXmv} \cite{nuxmv}, is an appropriate tool for our experiments. It supports both discrete and continuous domains, including real numbers $\mathbb{R}$ and unbounded integers $\mathbb{Z}$, and allows the use of propositional and temporal logics. 
\section{FANNet: Our Methodology for the Formal Analysis of Neural Network Properties} \label{PM}

Fig. \ref{meth} provides a comprehensive view of our FANNet methodology comprised of the following three key procedures.

\subsection{Behavior Extraction}
The NN testing samples $(X)$ and their true labels $(S_x)$ are available at the input. In this step, the weights $(w_1,w_2)$ and activations of the trained NN, which determine the network's architecture and functionality, are first translated into the \textit{SMV model} and the \textit{Temporal properties}, using the formal language of the model checker. Prior to the analysis with input noise, the correctness of the NN model without noise is ensured by comparing the model's calculated outputs $(OC)$ against $S_x$, i.e., \textit{P1} in Fig. \ref{meth}.

\subsection{Noise Tolerance Analysis}
The formal analysis for noise tolerance, 
proceeds as follows:
\begin{enumerate}[leftmargin=*]
    \item A large noise range for the analysis is initialized. 
    From this specified range, a \textit{unique} noise vector $(NV)$ is non-deterministically selected in each iteration.
    \item Noise is added \textit{relatively} to the input, and the input then transverses through the network layers.
   \item $OC_n$ is then computed and compared to $S_x$ of the input. Until counterexamples to $OC_n = S_x$ (i.e., \textit{P2} in Fig. \ref{meth}) are available, the noise is reduced iteratively. As soon as \textit{P2} becomes true, the noise tolerance of the NN is obtained.  
    \item The process is repeated for all inputs in the dataset.
\end{enumerate}

\subsection{Adversarial Noise Vector Extraction}
If $OC_n = S_x$ and the $NV$ is not already contained in $e$, then the $NV$ obtained from the generated counterexample is added to $e$, as shown by \textit{P3} in Fig. \ref{meth}. This ensures that $e$ is an array of unqiue noise patterns to which the NN is vulnerable.
\section{Case Study on Leukemia Diagnosis} \label{Result}

\subsection{Problem Description}
\begin{figure}[!t]
	\centering
	\includegraphics[width=\linewidth]{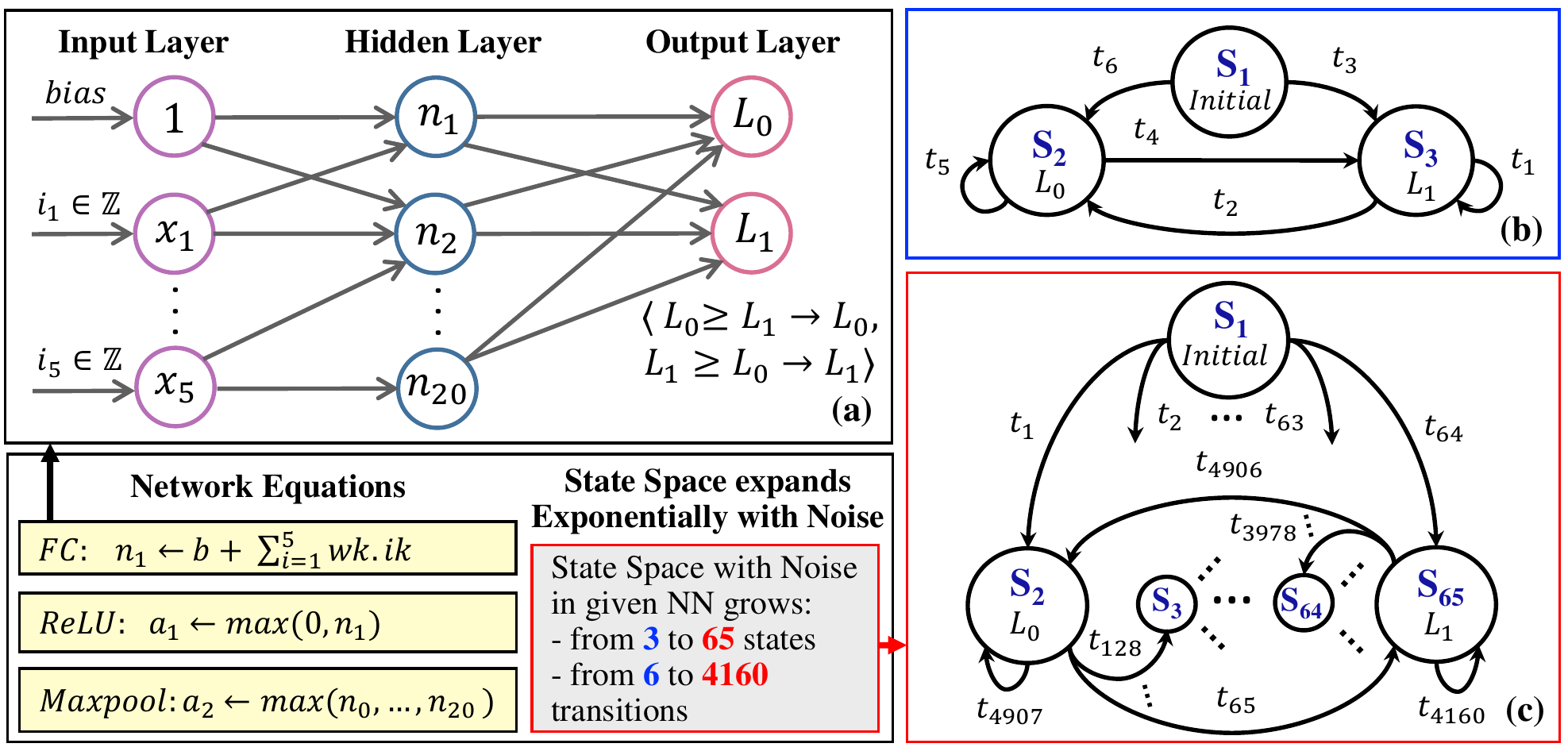}
	\vspace*{-5mm}
	\caption{(a) Feed-forward FC-NN trained for Leukemia Classification; (b) FSM for the given NN, in the absence of input noise; (c) FSM for NN with Noise Range $[0,1]\%$}
	\label{sp}
\end{figure}

\vspace{-1mm}
We considered a feed-forward FC neural network architecture as shown in Fig. \ref{sp}. It comprises of an input layer, one hidden layer and an output layer, with $6$, $20$ and $2$ nodes respectively. The activations used in the network are ReLU and Maxpooling. We trained the network\footnote{The network is trained using MATLAB with a learning rate of $0.5$ for the $40$ initial epochs, and a learning rate of $0.2$ for the remaining $40$ epochs. The training accuracy of the network is $100\%$ and the testing accuracy is $94.12\%$} to diagnose leukemia using the standard \textit{Leukemia} database \cite{dataset}, consisting of $38$ training samples and $34$ testing samples containing genetic attributes for Acute Lymphoblast Leukemia (ALL) and Acute Myeloid Leukemia (AML). In total, each data sample has $7129$ of these genetic attributes. From these, the top five most significant genes are picked as NN inputs using the Minimum Redundancy and Maximum Relevance (mRMR) feature selection method \cite{matlab-leukemia}.
\begin{figure*}
	\centering
	\includegraphics[width=\linewidth]{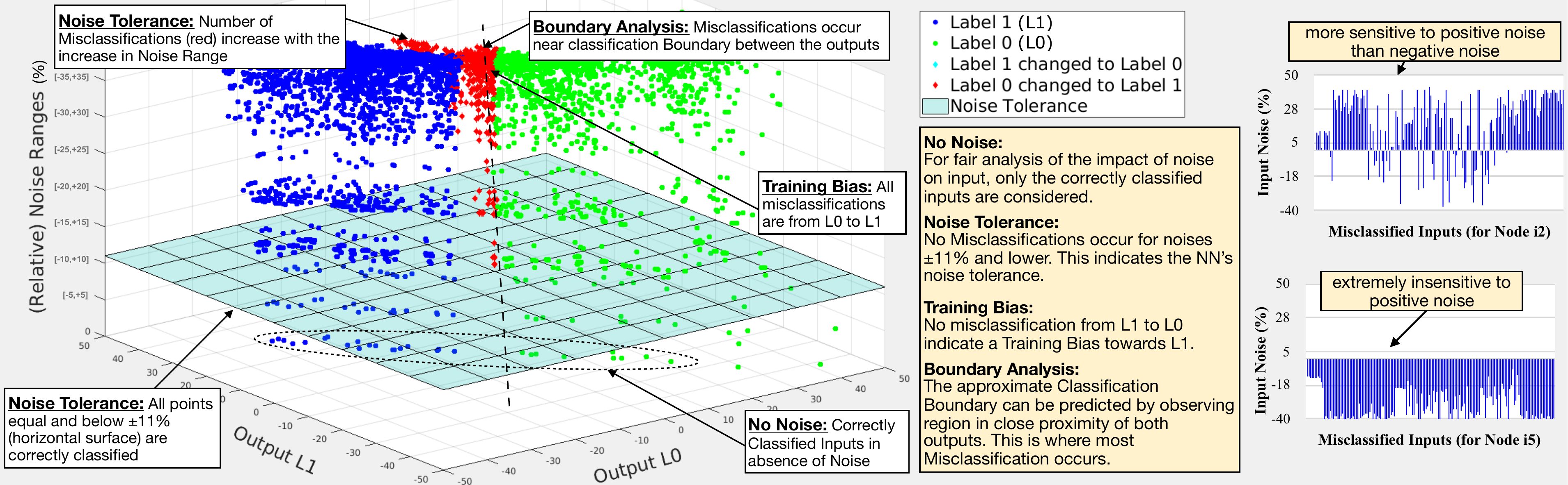}
	\vspace*{-5mm}
	\caption{Estimated Noise Tolerance, Impact of Noise on inputs closer to Classification Boundary, Effects of Training bias on NN Accuracy, and Sensitivity of specific Input Nodes}
	\vspace{-12pt}
	\label{results1}
\end{figure*}

\vspace{-1mm}
\subsection{FANNet Implementation}
\vspace{-2mm}
Varying noise ranges were input to the network, as explained in the previous section. The main goals of the analysis are: (a) to determine noise tolerance for the given network by gradually reducing the applied noise $\Delta x$ until no noise pattern that causes the true label $S_x$ to change can be found, and (b) to study network properties like training bias and input node sensitivity on the basis of the obtained counterexamples. It must also be highlighted that the objective of this work is not to exhaustively search for counterexamples, but rather to explore network properties on the basis of obtained counterexamples.
\vspace{-1mm}
\subsection{Formal Analysis of the Neural Network Properties}
\vspace*{-2mm}
We initiate the experiment using \textit{nuXmv} with a large input noise, and gradually reduce the noise until \textit{nuXmv} can verify the absence of any counterexamples for the given noise. The observations and analysis from our experiments, as illustrated in Fig. \ref{results1}, are as follows:

\subsubsection{\textbf{Noise Tolerance}}
For all the correctly classified inputs in the testing dataset, addition of a noise $\pm 11\%$ or less does not trigger misclassification for the given NN. Hence, assuming input noise to be the integer percentage values, the given NN has an estimated noise tolerance of $\pm 11\%$.

\subsubsection{\textbf{Classification Boundary Estimation}}
A few inputs among the dataset (i.e., inputs closer to the classification boundary) were observed to be highly susceptible to input noise. On the other hand, for other inputs, noise even as large as $50\%$ of the input did not trigger misclassification at the output. This knowledge can be used to estimate the network's classification boundary in the hyperspace.

\subsubsection{\textbf{Training Bias}}
Inputs with $S_x=L_0$ were observed as more likely to be misclassified than the inputs with $S_x=L_1$. On a closer inspection of the training dataset, it is observed that approximately $70\%$ of the data samples belong to the output class $L_1$ i.e., the training is biased towards the output $L_1$. This is corroborated by our formal analysis, where the misclassification of inputs with true output $L_0$ is more probable than the misclassification of inputs with true output $L_1$.

\subsubsection{\textbf{Input Node Sensitivity}}
No counterexamples were obtained with positive noise at input node $i_5$. Moreover, the counterexamples suggest more noise patterns with positive noise at input node $i_2$ than the other way around. The knowledge of the input node sensitivity, in some applications, could be exploited in the design of variable-precision data acquisition methodologies, where the resource-greedy measurements could be reserved for obtaining the sensitive inputs.

\section{Conclusion}\label{conclusion}
Input noise is known to trigger output misclassification in Neural Networks (NNs). Traditionally, the formal analysis of NNs focuses on checking network robustness for different $L^p-norm$ bounded noise. However, other network properties, i.e., network tolerance, training bias, and input node sensitivity, affecting NN's performance and are often neglected. To the best of our knowledge, this is the first model checking based formal analysis methodology that aims at investigating these properties affecting NNs. We use our methodology to estimate noise tolerance for a NN trained for Leukemia classification, to investigate training bias in the network due to the biased dataset, and to explore the sensitivity of the different genetic attributes presented by network's input nodes.
\section*{Acknowledgment}
This work was partially supported by Doctoral College Resilient Embedded Systems which is run jointly by TU Wien's Faculty of Informatics and FH-Technikum Wien, and partially supported by the Erasmus+ International Credit Mobility (KA107).



\bibliographystyle{IEEEtran}
\bibliography{References/Ref}


\end{document}